\documentclass[a4paper]{article}

\usepackage{INTERSPEECH2022}
\usepackage{url}
\usepackage{booktabs}
\usepackage{multirow}
\usepackage{subcaption}
\usepackage[section]{placeins}
\usepackage{adjustbox}

\title{Exploring Continuous Integrate-and-Fire for Adaptive \\ Simultaneous Speech Translation}
\name{Chih-Chiang Chang, Hung-yi Lee}
\address{
  National Taiwan University}
\email{r09922057@ntu.edu.tw, hungyilee@ntu.edu.tw}

\begin{document}

\maketitle
\begin{abstract}
 Simultaneous speech translation (SimulST) is a challenging task aiming to translate streaming speech before the complete input is observed. A SimulST system generally includes two components: the pre-decision that aggregates the speech information and the policy that decides to read or write. While recent works had proposed various strategies to improve the pre-decision, they mainly adopt the fixed wait-k policy, leaving the adaptive policies rarely explored. This paper proposes to model the adaptive policy by adapting the Continuous Integrate-and-Fire (CIF). Compared with monotonic multihead attention (MMA), our method has the advantage of simpler computation, superior quality at low latency, and better generalization to long utterances. We conduct experiments on the MuST-C V2 dataset and show the effectiveness of our approach.
\end{abstract}
\noindent\textbf{Index Terms}: simultaneous speech translation, streaming, continuous integrate-and-fire, online sequence-to-sequence model, end-to-end model

\section{Introduction}

Simultaneous translation is a task that performs translation in a streaming fashion. It requires producing the translation of partial input before the complete input is observed. It is challenging because it needs to take reordering between languages into account.
To achieve the best trade-off between latency and quality, a simultaneous translation system needs to decide whether to read more input or write a new token at each timestep. This decision can follow either a \textit{fixed policy}~\cite{ma2019stacl} or a \textit{flexible (adaptive) policy}~\cite{arivazhagan2019monotonic}.
Typically, a simultaneous machine translation (SimulMT) system is cascaded with a streaming speech recognition (ASR) to form practical applications such as international conferences.

End-to-end simultaneous speech translation (SimulST) aims to directly perform simultaneous translation on the speech input. Compared to cascaded SimulMT, end-to-end methods avoid error propagation~\cite{sperber2020speech,ruiz2014assessing} and are faster thanks to a unified model. However, it faces more challenges like acoustically ambiguous inputs or variable speech rates. Because speech input may be too fine-grained for policies to be learned, the pre-decision was introduced~\cite{ma2020simulmt}. The pre-decision segments the speech based on fixed chunks (\textit{fixed}) or word boundaries (\textit{flexible}), before read-write policies are applied. Most research on SimulST improves the speech encoding or the pre-decision, while adopting the fixed policy~\cite{nguyen21d_interspeech,ren2020simulspeech,ma2021streaming,chen2021direct,zeng2021realtrans,dong2021unist}. To our best knowledge, the only exceptions are the monotonic multihead attention (MMA) with pre-decision~\cite{ma2019monotonic,ma2020simulmt}, or the Cross Attention Augmented Transducer (CAAT)~\cite{liu2021cross}.

\begin{figure}[t!]
    \centering
    \includegraphics[width=\linewidth]{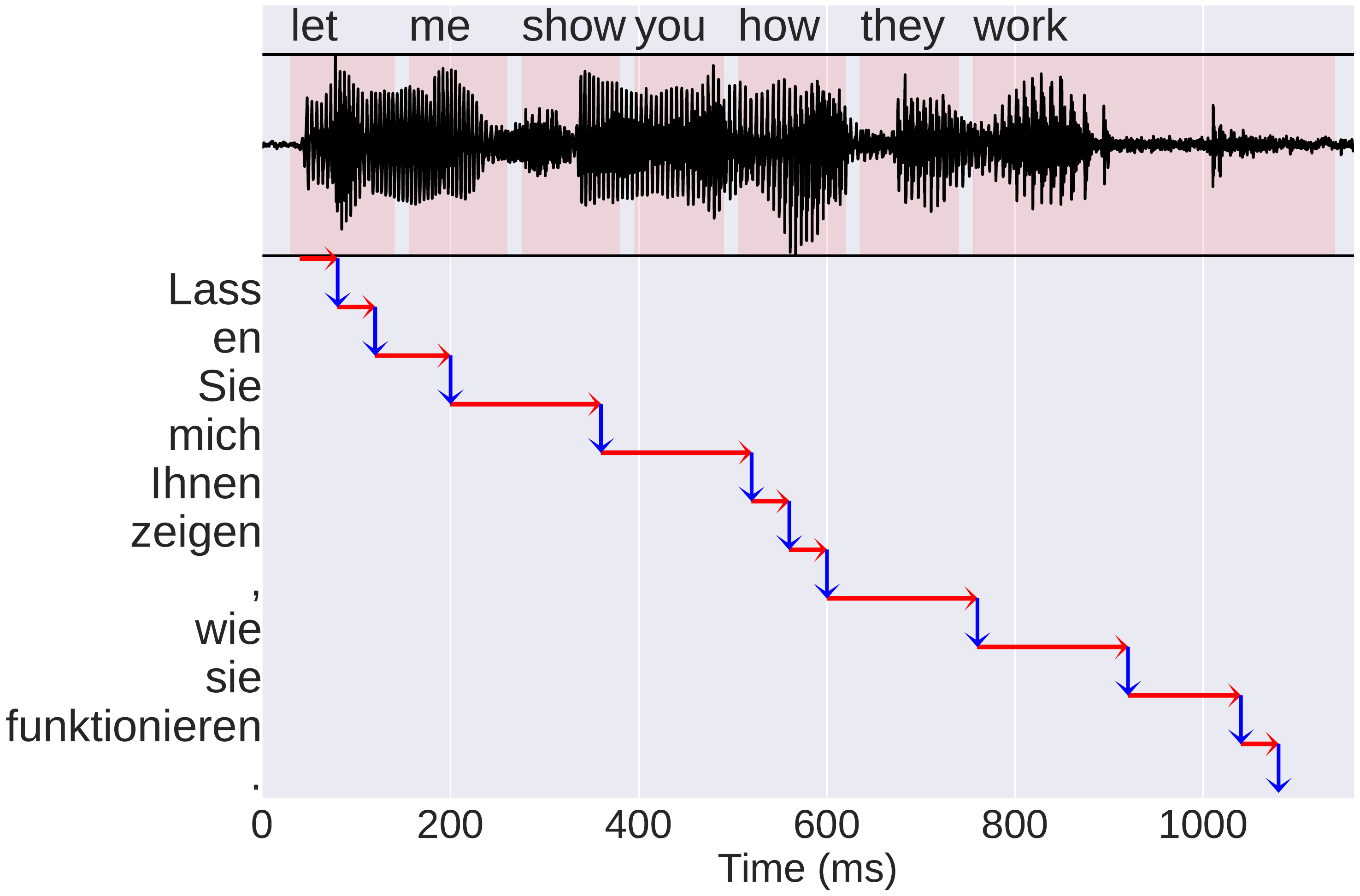}
    \caption{The illustration of the adaptive policy learned by the CIF module. From top to bottom: the transcription, the speech with force alignment and the translation read-write policy.
    Red arrows are CIF integrations and correspond to read actions. Blue arrows are CIF firings and correspond to write actions.
    }
    \label{fig:policy}
\end{figure}

In this paper, we explore the Continuous Integrate-and-Fire (CIF)~\cite{dong2020cif} for another adaptive policy (Figure~\ref{fig:policy}). Similar to connectionist temporal classification (CTC)~\cite{graves2006connectionist} or
the MMA,
the CIF is also a monotonic alignment method.
It has several advantages over the MMA.
During training, the MMA marginalizes out all alignments at \textit{each} decoder layer. Besides, gradient clipping, denominator clipping and log-space conversion\footnote{interestingly, the official MMA implementation computes cumulative product in the linear-space instead of log-space.} are essential to alleviate numerical instability~\cite{raffel2017online}.
The CIF, on the other hand, only computes a single alignment using addition and weighted sum, and is thus faster and more stable.
Furthermore, in our experiments, the CIF not only exhibits better quality in low latency, but also generalizes better to long utterances. 
We argue that being able to deal with longer utterances is an important advantage because, in practice, the input speech can be arbitrarily long, and sentence boundary is often unavailable. On the other hand, the quality at low latency is more critical for streaming because a second-pass model or even an offline model may be more appropriate for higher latency.



To adapt the CIF to SimulST, we proposed two modifications. The first is extending the quantity loss~\cite{dong2020cif} to a token-level objective. It leverages the forced alignment from an auxiliary CTC prediction head. The other is allowing the decoder to attend to past CIF integration states for the sake of explicit reordering.
We test our result on MuST-C V2~\cite{CATTONI2021101155} test set, as well as three synthetic long utterance set constructed from it.
Our contributions are summarized below:
\begin{itemize}
    \setlength\itemsep{0em}
    \item We proposed a new approach to model the adaptive policy in SimulST.
    \item The proposed method is fast, stable, performs better at low latency, and generalizes better to long utterances.
\end{itemize}

\section{Related Works}
\subsection{Continuous Integrate-and-Fire}
The CIF was proposed for ASR as an online model to learn the precise acoustic boundaries~\cite{dong2020cif}.
It is faster and more robust to long or noisy utterances during inference~\cite{dong2020comparison} than the standard Transformer~\cite{vaswani2017attention}.
Since it integrates longer sequences into shorter ones, the CIF can bridge the gap between pre-trained acoustic and linguistic representations~\cite{yi2021efficiently}.
Applied to SimulST, it can act as pre-decision to locate acoustic boundaries, so that fixed policies can be applied~\cite{dong2021unist}.
Our work differs from these works in that we leverage CIF to learn the adaptive policy of SimulST. We do not use the transcription when training translation models.


\subsection{End-to-End Simultaneous Speech Translation}

In recent years, several works had focused on improving the speech encoding and the pre-decision process. \cite{nguyen21d_interspeech} studied the encoding and segmentation strategies for ULSTM encoder. SimulSpeech~\cite{ren2020simulspeech} used CTC for pre-decision, and used multi-task learning and knowledge distillation to improve SimulST. \cite{ma2020simulmt} introduced fixed pre-decision and flexible pre-decision. \cite{ma2021streaming} explored block processing for the streaming encoder. \cite{chen2021direct} used an ASR module as flexible pre-decision. RealTranS~\cite{zeng2021realtrans} used CTC for pre-decision, and semantic encoder to improve encoding. \cite{dong2021unist} used self-supervised speech encoder, CIF as pre-decision, and a semantic encoder as well. All these works adopted the \textit{wait-$k$} or similar fixed policies.

On the other hand, adaptive policies have rarely been explored for speech. \cite{ma2020simulmt} explored MMA for its decoder. CAAT~\cite{liu2021cross} augmented the RNN-T~\cite{graves2012sequence} with cross-attention to have a more complex joiner capable of reordering.
Our work introduces a new adaptive policy learned using the CIF. We compare our results with that of the MMA.
\begin{figure}[t!]
    \centering
    \includegraphics[width=1.0\linewidth]{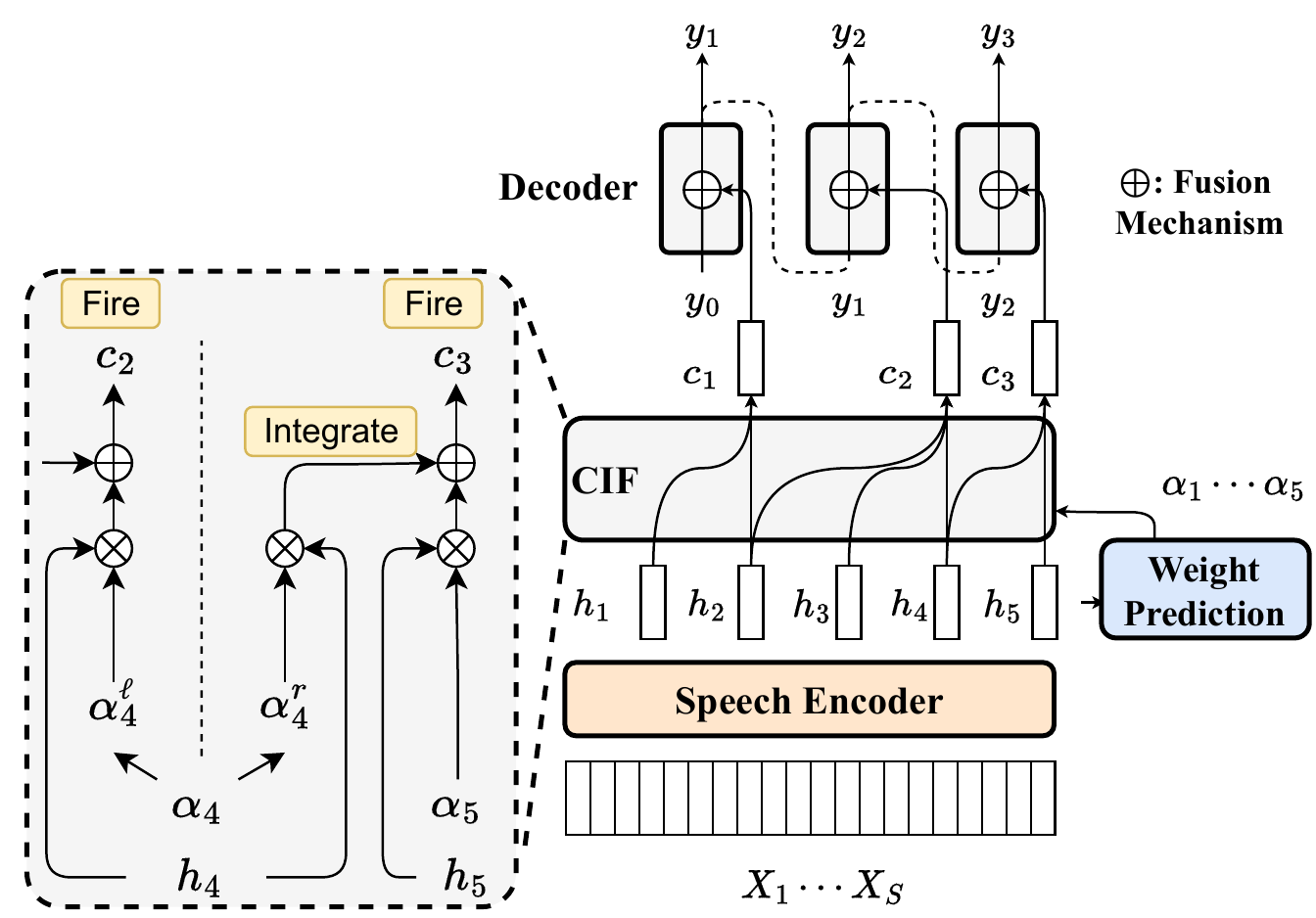}
    \caption{The architecture of CIF-based model. 
    It consists of a speech encoder, a weight prediction, CIF, and a decoder. The speech encoder transforms the input speech into contextualized features. 
    The weight prediction predicts weights used by CIF.
    The CIF learns the monotonic alignment that corresponds to a read-write policy.
    The integrated embeddings $\langle c_i\rangle$ are fused with decoder states via one of the fusion mechanisms.
    }
    \label{fig:arch}
\end{figure}

\section{Method}

Figure~\ref{fig:arch} demonstrates the overall architecture of our model. 
We describe each component below.

\subsection{Speech Encoder}
The speech encoder takes a sequence of speech features $X_1, \cdots ,X_{S}$ as input, and produces the encoder states $h_1,h_2,\cdots,h_U$.
It consists of a feature extractor, a positional encoding and an Emformer~\cite{shi2021emformer}.
We use the feature extractor from the fairseq-S2T~\cite{wang2020fairseq}.
For positional encoding, we use a temporal convolution to capture relative positional information~\cite{mohamed2019transformers}.
The Emformer is a block processing Transformer suitable for streaming. Each block has access to its main context, left context, right context and a memory bank. 
During training, it uses a block diagonal attention mask to remove future context; it hard copies the right context for each block to prevent the look-ahead context leakage~\cite{shi2021emformer}. 
During inference, it encodes the input block by block with overlapping context.

\subsection{Weight Prediction and CIF Mechanism}
\newcommand{\Queh}{\mathcal{Q}_{h}}
\newcommand{\Quea}{\mathcal{Q}_{\alpha}}

In this work, the CIF learns when to output a translation token.
Suppose the target sequence is $y_1,y_2,\cdots,y_T$.
For each encoder state $h_j, \forall j\in[1,U]$, we first use a weight prediction network to predict $\alpha_j$. Then, the CIF caches $h_j$ and $\alpha_j$ in the accumulation queues $\Queh$ and $\Quea$, respectively. If the current accumulated weights in $\Quea$ is below the threshold $\beta$, then the CIF simply proceeds to the next encoder step $j+1$. Otherwise, it needs to trigger the integrate and fire operation. 
The CIF divides $\alpha_j$ into $\alpha_j^\ell$ and $\alpha_j^r$. 
The $\alpha_j^\ell$ and the other weights in $\Quea$ will sum to $\beta$, so the CIF will integrate the vectors in $\Queh$ using weighted sum with these weights, in order to produce the integrated embedding $c_i$. The remaining weight $\alpha_j^r$ and and a copy of $h_j$ will be kept in $\Quea$ and $\Queh$ for the next integration. Finally, $c_i$ is send to the decoder, which is called a firing. The above process is performed till the end (when $j=U$).


The weight prediction network is composed of a temporal convolution, followed by layer normalization~\cite{ba2016layer}, 
a GELU~\cite{hendrycks2016gaussian},
dropout, a fully connected layer with one output and a sigmoid. 
We use stop gradient on the input of weight prediction to improve generalization.
The intuition is that the encoder should focus on encoding speech, and weight prediction should be done by a separate module.

During training, the CIF won't fire exactly $T$ times. 
This brings difficulties to the cross-entropy objective.
To make sure the sequences $\langle c_i \rangle$, $\langle y_i \rangle$ match in length, two strategies were proposed by~\cite{dong2020cif}. First, before CIF, the scaling strategy normalizes the weight of each timestep, so that they sum to $\beta T$:
\begin{equation}\textstyle
    \alpha_j' = \alpha_j \cdot \beta T / \sum\nolimits_{k=1}^U\alpha_k 
\end{equation}
Next, a squared L2 quantity loss encourages the number of fires to be around $T$:
\begin{equation}\textstyle
    \mathcal{L}_{qua} = \left\|T - \sum\nolimits_{j=1}^U\alpha_j / \beta \right\|_2^2 \label{eq:l_qua_sum}
\end{equation}
These strategies are only used during training but not inference.

\subsection{Decoder and Fusion Mechanism}
Our decoder follows the standard autoregressive Transformer decoder, except for the cross-attention which is replaced by a fusion mechanism. We design two fusion mechanisms: one is position-wise; another is with infinite lookback attention (ILA)~\cite{arivazhagan2019monotonic}. 

\subsubsection{Position-wise Fusion}
Because the number of integrated embeddings $c_i$ is the same as decoder length $T$, we replace the cross-attention in each layer with a position-wise fusion described below:
\begin{equation}
    \mathrm{Fusion}(c_i, s_i) = W_o f_{act}(W_s c_i+ W_t s_i+b),
\end{equation}
where $s_i$ is the decoder hidden state after self-attention. $W_o, W_s, W_t$ are trainable weight matrices and $b$ is a bias vector. $f_{act}$ is GELU here. While fusion operations are the same across different positions, they use different parameters from layer to layer. We refer to this model as CIF-P.

\subsubsection{Infinite Lookback Fusion}
The ILA improves monotonic models for SimulST thanks to its ability to explicitly reorder~\cite{arivazhagan2019monotonic,liu2021cross}. Thus, we also explore having the decoder attend to past integrated embeddings. In this variant, the cross-attention is retained, but the $i$-th deocder state can only attend to integrated embedding $c_k$ if $k\leq i$.
We refer to this model as CIF-IL.


\subsection{Token-level Quantity Loss}
Preliminary experiments show that directly applying the CIF-P to SimulST has limitations. 
In particular, the validation $\mathcal{L}_{qua}$ in $\eqref{eq:l_qua_sum}$ increases while the training value decreases.
We suspect this is because optimizing $\mathcal{L}_{qua}$ at the sequence-level provides a weak training signal for alignment. Since the CTC loss often accompanies the CIF~\cite{dong2020cif,dong2020comparison,yi2021efficiently}, this inspired us to leverage its alignment for token-level objective. During training, we use the CTC forward-backward algorithm to obtain the forced alignment. The sequence $a_1,a_2,\cdots,a_U\in\{\phi, y_1,\cdots,y_T\}$ monotonically align each encoder position to a target position or a blank ($\phi$). Then, we find boundaries by looking for $j$ such that $a_j\not=\phi$ and $a_j\not=a_{j+1}$. For each boundary position $j$, we further define its target length as $t_j=i$, where $i$ is the index of its aligned token $y_i$ (i.e. $a_j=y_i$).
Finally, we define a token-level quantity loss:
\begin{align}
    \mathcal{L}_{qua}(j) &= \begin{cases}
    \left\|t_j - \frac{1}{\beta}\sum_{k=1}^{j}\alpha_k \right\|_2^2, &\text{if $j$ is boundary} \\
    0, &\text{otherwise}
    \end{cases} \\
    \mathcal{L}_{qua} &= \textstyle\frac{1}{T}\sum\nolimits_{j=1}^{U}  \mathcal{L}_{qua}(j) \label{eq:l_qua_align}
\end{align}
The CIF-IL has access to past integrated embeddings, so the alignment is easier to learn. We observe little difference between using \eqref{eq:l_qua_align} and \eqref{eq:l_qua_sum} when training CIF-IL, so \eqref{eq:l_qua_sum} is used.

\subsection{Latency Loss}
To provide a latency-quality trade off, we add latency training with the Differentiable Average Lagging (DAL)~\cite{arivazhagan2019monotonic}. To do so, we first need to define the expected delay $d(y_i)$ of a target token $y_i$, which is the expected source length processed before $y_i$ is predicted. Suppose that $y_i$ is predicted using $c_i$, and that: 
\begin{equation}\textstyle
    c_{i} = \alpha_m^r h_{m} + \alpha_{m+1}h_{m+1}+\cdots+\alpha_{j-1}h_{j-1}+\alpha_j^l h_{j}.
\end{equation}
Then we define:
\begin{align}\textstyle
    d(y_i) = \frac{1}{\beta}(
        \alpha_m^r m + \alpha_{m+1}(m+1)+\cdots \nonumber\\
        +\alpha_{j-1}(j-1)+\alpha_j^l j
    ). 
\end{align}
The latency loss $\mathcal{L}_{lat}$ is the DAL computed with the expected delays. Due to space limitation, we refer the readers to~\cite{arivazhagan2019monotonic} for details on latency training.

\subsection{Final objective}
The final objective is described by:
\begin{equation}
    \mathcal{L} = \mathcal{L}_{ce} + \lambda_{ctc}\mathcal{L}_{ctc} + \lambda_{qua}\mathcal{L}_{qua} + \lambda_{lat}\mathcal{L}_{lat},
\end{equation}
where $\mathcal{L}_{ce}$ is the standard cross-entropy loss. $\lambda_{ctc}=0.3$, $\lambda_{qua}=1.0$ for all experiments, and $\lambda_{lat}\in\{0, 0.5, 1.0, 1.5, 2.0\}$ for both CIF-P and CIF-IL.
\section{Experimental Setup}
\label{sec:exp}
We conduct experiments on the English-German (En-De) portion of MuST-C V2~\cite{CATTONI2021101155}.
The V2 released by IWSLT 2021~\cite{anastasopoulos2021findings} contains 435 hours of speech and 249k sentence pairs.
We use the pre-processing pipeline from fairseq-S2T~\cite{wang2020fairseq}. Input is transformed to 80 dimensional log-mel filter bank features using 25ms window and 10ms shift. We keep training examples within 5 to 3000 frames. SpecAugment~\cite{park19e_interspeech} with the LB policy is applied.
SentencePiece~\cite{kudo2018sentencepiece} is used to generate subword vocabulary with 4096 tokens for each language pairs. We tune the performance using the dev set, and report results on the tst-COMMON set.

The feature extractor has 2 convolution layers with kernel size 5, stride 2, and hidden dimension 256. The positional encoding has kernel size 64, group size 16 and stride 1. The Emformer encoder has 12 layers and the decoder has 6 layers. The models have 4 attention heads, $d_{model}=256$ and $d_{FFN}=2048$. Dropout is set to 0.1 for activations and attentions, and 0.3 for others.
We follow~\cite{ma2021streaming} and set the Emformer main context to 640 ms, right context to 320 ms, memory bank to 5. We increase the left context to 1280 ms following~\cite{shi2021emformer}. The convolution in the CIF weight prediction has kernel size 3 and stride 1. The CIF threshold $\beta=1$ during training, and is individually tuned for each model on the dev set during inference. Tail handling~\cite{dong2020cif} uses $\beta/2$.
To enable streaming, all the convolution layers are \textbf{causal}. The 320 ms right context of the Enformer is the only look-ahead.
The MMA models use fixed pre-decision ratio 8 (i.e. 320 ms).
Latency weight is $\lambda_{var}\in\{0.02, 0.04, 0.06, 0.1, 0.2, 0.4\}$, and $\lambda_{avg}=0$ for MMA-H and $\lambda_{avg}=\lambda_{var}$ for MMA-IL.
Label smoothing~\cite{szegedy2016rethinking} of 0.1 is applied.
We use inverse square root schedule and 4000 warm-up steps for the optimizer. The max learning rate is 1e-4 for MMA-H\footnote{training diverges with 1e-3 due to instability.}, and 1e-3 for others.
We use up to 4 V100 GPUs and gradient accumulation to achieve an effective batch size of 160k frames.
Gradient norm are clipped at 10, and weight decay is set to 1e-6.
We follow~\cite{ma2020simulmt} to first train without latency loss for 150K steps, then finetune with various latency weights for another 50k steps.
We average the 5 checkpoints with best latency for evaluation.
To ease optimization, all models use pre-trained encoder and sequence-level knowledge distillation (Seq-KD)~\cite{kim2016sequence}. Specifically, the encoders are initialized from an ASR model pre-trained under the joint CTC and cross-entropy loss for 300k steps.
For Seq-KD we first train a NMT model on the sentence pairs, then we use beam search with width $5$ to decode the Seq-KD set, which is used as the new training data for ST.

We use SimulEval~\cite{ma2020simuleval} to compute speech versions of widely adopted latency metrics including Average Proportion (AP)~\cite{cho2016can}, Average Lagging (AL)~\cite{ma2019stacl} and DAL~\cite{cherry2019thinking}. We report case-sensitive detokenized BLEU using SacreBLEU~\cite{post2018call}, the signature is provided\footnote{BLEU+case.mixed+numrefs.1+smooth.exp+tok.13a+version.1.5.1}. Our implementation is available\footnote{\url{https://github.com/George0828Zhang/simulst}}.

\begin{figure*}[!h]
    \centering
    
    \subcaptionbox{DAL (ms)\label{fig:dal_tst}}{\includegraphics[width=0.315\textwidth]{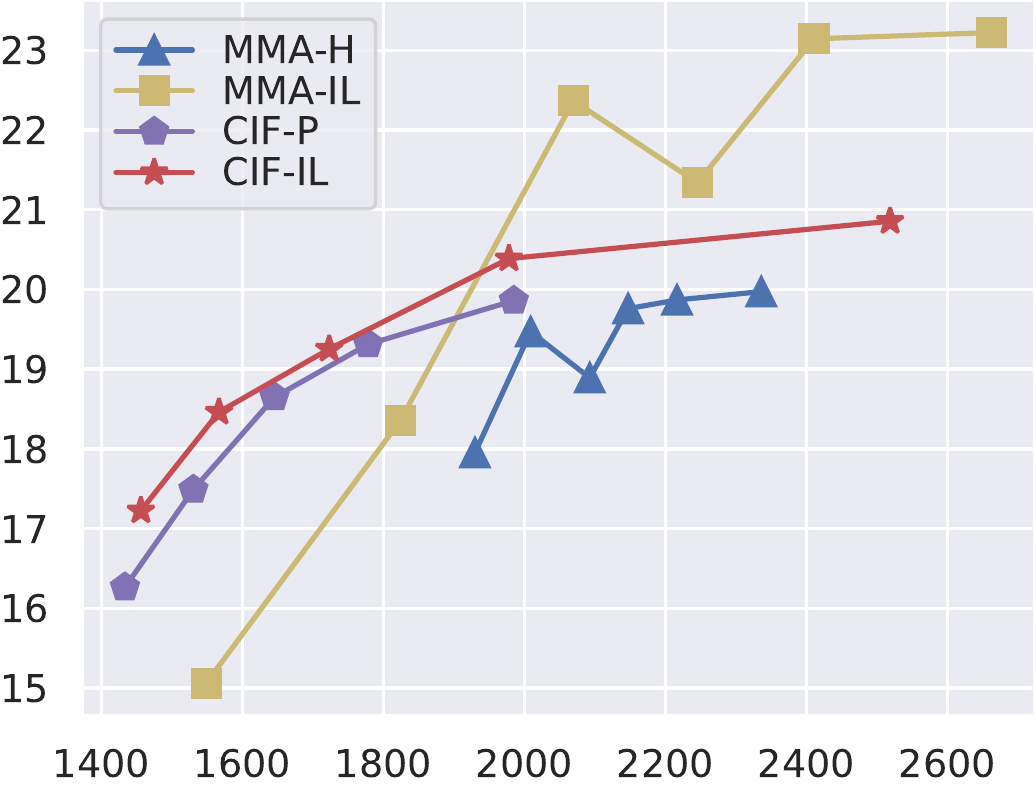}}\quad
    \subcaptionbox{AP\label{fig:al_dev}}{\includegraphics[width=0.315\textwidth]{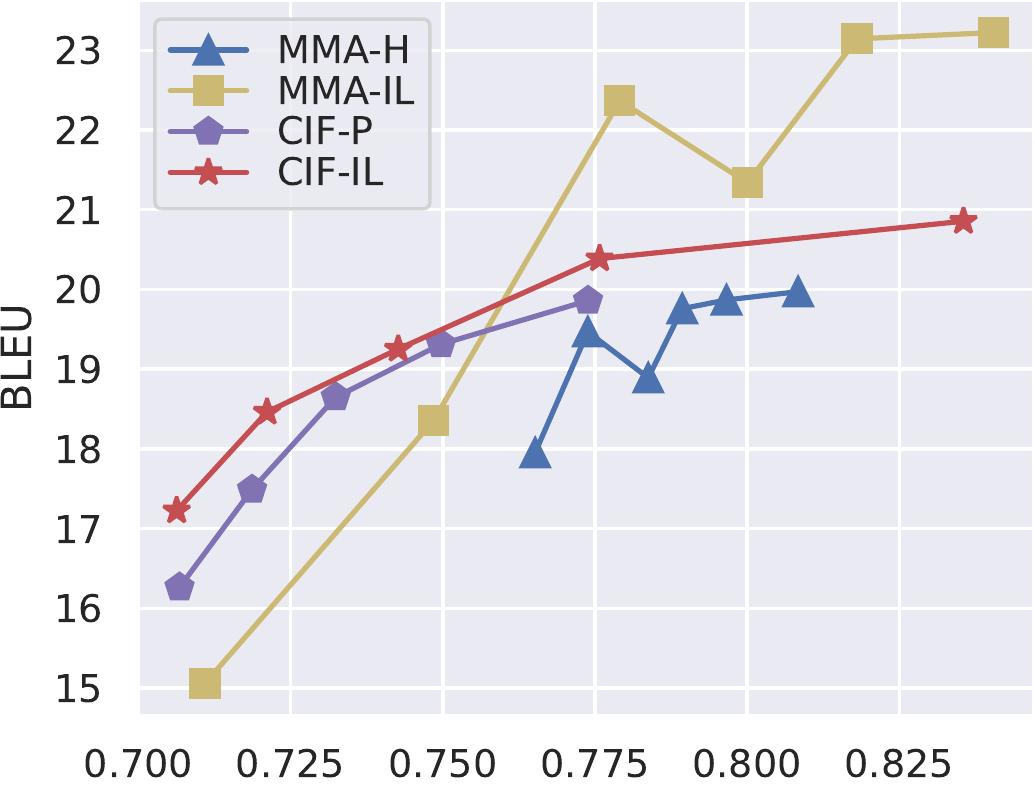}}\quad
    \subcaptionbox{AL (ms)\label{fig:dal_dev}}{\includegraphics[width=0.33\textwidth]{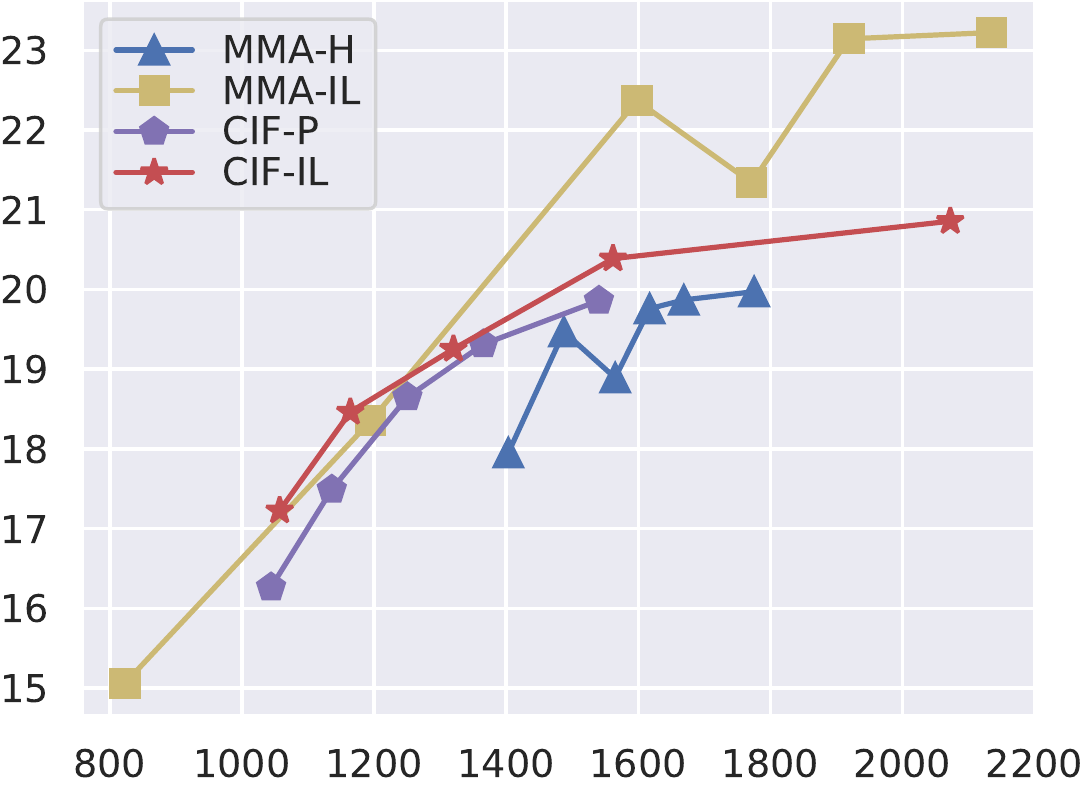}}
    
    \vspace{-3mm}
    \caption{Latency-quality trade off on tst-COMMON. The 5 dots for CIF are $\lambda_{lat}\in\{0, 0.5, 1.0, 1.5, 2.0\}$. The 6 dots for MMA are $\lambda_{var}\in\{0.02, 0.04, 0.06, 0.1, 0.2, 0.4\}$.}
    \label{fig:lateny-quality}
\end{figure*}

\begin{table}[]
\caption{Performance comparison on long utterance sets with $L\in\{0, 20, 40, 60\}$. $L=0$ is the original set. The top 4 rows are settings with best BLEU. The bottom 4 are settings with similar BLEU. DAL is measured in seconds here.}
\centering
\adjustbox{max width=\linewidth}{%
\setlength{\tabcolsep}{1pt}
\renewcommand{\arraystretch}{1}
\begin{tabular}{c|cc|cc|cc|cc}
\hline
\textbf{} & \multicolumn{2}{c|}{\textbf{L=0}} & \multicolumn{2}{c|}{\textbf{L=20}} & \multicolumn{2}{c|}{\textbf{L=40}} & \multicolumn{2}{c}{\textbf{L=60}} \\ \hline
\textbf{Model} & \textbf{BLEU} & \textbf{DAL} & \textbf{BLEU} & \textbf{DAL} & \textbf{BLEU} & \textbf{DAL} & \textbf{BLEU} & \textbf{DAL} \\ \hline
{MMA-H} & 19.97 & 2.34 & 11.62 & 4.23 & 4.78 & 10.16 & 2.57 & 19.09 \\
{MMA-IL} & 23.22 & 2.66 & 19.55 & 7.81 & 13.26 & 18.87 & 3.41 & 35.37 \\
{CIF-P} & 19.86 & 1.99 & 18.28 & 2.92 & 16.39 & 3.39 & 14.76 & 3.95 \\
{CIF-IL} & 20.86 & 2.52 & 18.16 & 3.85 & 13.53 & 4.54 & 10.62 & 5.50 \\ \hline
{MMA-H} & 18.89 & 2.09 & 13.04 & 3.60 & 5.28 & 8.62 & 3.36 & 15.17 \\
{MMA-IL} & 18.35 & 1.82 & 17.11 & 4.63 & 10.02 & 12.09 & 5.89 & 21.64 \\
{CIF-P} & 18.65 & 1.65 & 17.43 & 2.50 & 16.02 & 2.99 & 14.78 & 3.52 \\
{CIF-IL} & 18.46 & 1.57 & 17.17 & 2.60 & 13.24 & 3.25 & 10.79 & 4.00 \\ \hline
\end{tabular}}
\label{tab:long-utt}
\vspace{-3mm}
\end{table}

\begin{table}[h!]
\caption{Time comparison. total: the total time from the second epoch to 50k steps. step: the wall clock time of each step. The DAL-CA is computed on an Intel Xeon E5-2620 CPU using a single core, and is averaged over 3 evaluations.}
\centering
\adjustbox{max width=\linewidth}{%
\setlength{\tabcolsep}{1.5pt}
\begin{tabular}{c|cc|ccc}
\hline
 & \multicolumn{2}{c|}{\textbf{train}} & \multicolumn{3}{c}{\textbf{inference}} \\ \hline
\textbf{Method} & \multicolumn{1}{c|}{\textbf{total (hr)}} & \textbf{step (ms)} & \multicolumn{1}{c|}{\textbf{DAL (ms)}} & \multicolumn{1}{c|}{\textbf{DAL-CA (ms)}} & \textbf{$\Delta$} \\ \hline
MMA-H & \multicolumn{1}{c|}{13.9} & 50 & \multicolumn{1}{c|}{2006} & \multicolumn{1}{c|}{2981} & 975 \\
MMA-IL & \multicolumn{1}{c|}{18.3} & 66 & \multicolumn{1}{c|}{2068} & \multicolumn{1}{c|}{3158} & 1089 \\
CIF-P & \multicolumn{1}{c|}{10.1} & 36 & \multicolumn{1}{c|}{1983} & \multicolumn{1}{c|}{2770} & \textbf{787} \\
CIF-IL & \multicolumn{1}{c|}{\textbf{9.1}} & \textbf{32} & \multicolumn{1}{c|}{1978} & \multicolumn{1}{c|}{2850} & 872 \\ \hline
\end{tabular}
}
\label{tab:train-speed}
\vspace{-3mm}
\end{table}

\section{Results}

\subsection{Comparison on the original test set}

Figure~\ref{fig:lateny-quality} shows the result on tst-COMMON. First, the performance of MMA-IL at low latency appear better with AL. It is shown by examples that AL may be exploited by adaptive policy~\cite{cherry2019thinking}. Therefore, we prefer DAL over AL for our analysis.
Interestingly, the MMA-H and the MMA-IL have drastically different performance in SimulST, despite being comparable in SimulMT~\cite{ma2019monotonic}. This may be due to the variable speech rate that is better handled by the ILA. Compared with MMA-H, both CIF-P and CIF-IL have lower latency at the same quality, and CIF-IL can achieve better quality. 
Compared with MMA-IL, proposed models has superior quality at lower latency (DAL$<$2000)\footnote{Low latency is technically AL$\leq$1000~\cite{anastasopoulos2021findings}. But our latency can be reduced by a smaller Emformer block size, to fit this definition.}, but falls short at higher latency.
This shows that the CIF is a better choice in medium-low latency, while MMA-IL is better if high latency is tolerable.


Figure~\ref{fig:policy} is an example from tst-COMMON that demonstrates the read-write policy from CIF-P. We observe that the policy is close to ideal as it is near the diagonal. Notably, the model learns to speed up easier predictions like the punctuation ``,'' or the subword ``en'' right after ``Lass'' to complete the word ``Lassen'' (the translation of ``let''). Besides, the model delays its prediction when translating ``show you'' into ``Ihnen zeigen'', which requires reordering.

\subsection{Comparison on long utterances}
We concatenate utterances from the same TED talk in the original order, so the content remain the same as the original test set. We make sure that each resulting utterance is at least $L$ seconds.
For each method, we evaluate the setting with the best BLEU or with a BLEU closest to 18.5. 
From Table~\ref{tab:long-utt} we see that for MMA, both the latency and quality deteriorate rapidly as the utterance length increases. 
When the input is a minutes long, the MMA-IL's latency is over 21 seconds and BLEU is less than 6.
In contrast, the latency and quality of CIF decline more gently. Particularly, CIF-P is the most robust, remaining at about 14.7 BLEU in the longest setting.
For MMA, multiple heads operate independently, and there are no mechanism preventing heads from being stuck at early timesteps.
For MMA-IL and CIF-IL, relying on the ILA have the risk of distracting the decoder with earlier, irrelevant inputs.
All these problems manifest as the input speech duration increases.


\subsection{Comparison on computation time}
We compare the training and inference time of models with similar DAL in Table~\ref{tab:train-speed}. The computation per token ($\Delta$) is the difference between the DAL and the computation aware DAL (DAL-CA)~\cite{ma2020simulmt}. 
Evidently, the CIF take much less time to train.
During inference, the CIF also have an advantage of up to 302 ms per token over the MMA. This advantage may further increase on computation-limited edge devices.
\section{Conclusion}

We applied the CIF mechanism to the SimulST task. We show that despite its simplicity, the CIF achieved decent latency-quality trade off. We verified that the CIF speeds up easier predictions and delays harder ones. Compared with MMA, CIF has the advantages of faster computation, superior quality at low latency, and better generalization to long utterances. We recommend future work on SimulST to take a closer look at long utterances as it is closer to practical application.

\bibliographystyle{IEEEtran}
\clearpage
\bibliography{mybib}


\end{document}